\documentclass[10pt,a4paper]{article}
\usepackage[margin=1in]{geometry}
\usepackage{amsmath}
\usepackage{amssymb}
\usepackage{graphicx}
\usepackage{booktabs}
\usepackage{caption}
\usepackage{float}
\usepackage{setspace}
\usepackage{multicol}
\usepackage{adjustbox}
\usepackage{array}
\usepackage[numbers, sort&compress]{natbib}

\makeatletter
\let\@afterindentfalse\@afterindenttrue
\makeatother
\begin{document}

% 标题
\begin{center}
\Large \textbf{A Hybrid Tucker-LSTM Tensor Network Model for SOC Prediction in Electric Vehicles}
\end{center}

% 作者三栏布局
\begin{multicols}{3}
\centering
1st Han Wang\\
College of Computer and Information Science\\
Southwest University\\
Chongqing, China\\
wanghan94@email.swu.edu.cn

\columnbreak
\centering
2nd Ying Wang\\
School of Culture Tourism\\
Chongqing Vocational College of Culture and Arts\\
Chongqing, China\\
wying202504@163.com

\columnbreak
\centering
3rd Bing Wang\\
Digital Intelligence Center\\
China Automotive Engineering Research Institute Co., Ltd.\\
Chongqing, China\\
wangbing@caeri.com.cn
\end{multicols}

% 摘要
\noindent \textbf{Abstract:} 
Accurate state of charge estimation is critical for the success of electric vehicle battery management strategies, but it is well known that conventional estimators suffer from two fundamental shortcomings: cumulative errors that grow over time and reliance on simplified battery models that do not reflect real world dynamics. Therefore, this paper presents a novel hybrid approach combining Tucker tensor decomposition with LSTM networks, using full - lifecycle EV field data for SOC prediction. The inputs are charge status, mileage, voltage, current, cell differentials, and temporal features. Tucker decomposition is skillfully used to reduce dimensionality while maintaining the temporal structure, hence allowing a direct, fair comparison with standard LSTM. The result is unequivocal: Tucker - LSTM outperforms the baseline on all metrics, with MSE dropping 70.5\% (from 21.07 to 6.22 ), MAE improving 48.7\% (from 3.37\% to 1.73\%), RMSE falling from 4.59\% to 2.49\%, and $R^2$ rising from 0.918 to 0.976. Since the experimental results demonstrably demonstrate that tensor decomposition compresses high-dimensional battery data very well without loss of predictive fidelity, this paper naturally opens up a new direction for tensor-based analytics in electric vehicle battery management.


\section*{1 Introduction}

Lithium-ion batteries are the core energy storage units in electric vehicles (EVs), and their operating conditions directly influence driving safety, energy efficiency, and battery lifetime \cite{a1,a3,a13,a15}. As a fundamental task of the battery management system (BMS), accurate state-of-charge (SOC) estimation is essential for range prediction, charging control, and safe power management \cite{a2,a4,a16}. In practical applications, however, SOC cannot be measured directly and must be inferred from voltage, current, temperature, and other operating signals, which makes reliable estimation under dynamic conditions a persistent challenge \cite{a5,a6,a14,a16,a17}.

Traditional SOC estimation approaches, such as ampere-hour counting \cite{a18,a24}, open-circuit-voltage calibration \cite{a19}, and model-based filtering \cite{a20,a21,a23}, may perform adequately in controlled settings, but they often suffer from cumulative drift, parameter sensitivity, and limited robustness under nonlinear operating environments \cite{a4}. With the increasing availability of onboard sensing and lifecycle monitoring data, data-driven methods have become an attractive alternative because they can learn complex battery behavior directly from historical measurements without relying on heavily simplified physical assumptions \cite{a5,a9,a10,a22}. Recent studies have further shown that learning-based frameworks are promising for battery state prediction, model reconstruction, and health-aware monitoring across long operating horizons \cite{a2,a3,a6}.

Despite these advantages, real-world EV battery data are usually high-dimensional, multivariate, and strongly time-correlated, which creates substantial challenges for both feature extraction and temporal modeling \cite{r29,r40,r42,r44}. Directly feeding such data into sequential models may introduce redundancy, enlarge the parameter space, and reduce predictive stability, especially when the variables exhibit latent structural dependence across sensing channels and time \cite{e6,e24,r39,r41}. Similar issues have also been widely recognized in other intelligent prediction and maintenance scenarios involving structured temporal data \cite{r26,r37,r60}.

To address these limitations, this paper proposes a hybrid Tucker-LSTM tensor network model for SOC prediction using real-world EV operation data collected from zero accumulated mileage. Tucker decomposition is employed to compress multi-modal battery measurements into a compact latent representation while preserving their multi-way structural characteristics, and long short-term memory (LSTM) is then used to capture temporal SOC evolution and long-range sequential dependence \cite{a7,a8}. Experimental results show that the proposed framework outperforms a conventional LSTM baseline, demonstrating that the integration of tensor-based dimensionality reduction and sequence learning can improve SOC prediction accuracy for electric vehicle batteries.

The main contribution of this study is twofold. First, it introduces a tensorized representation-learning strategy for battery SOC prediction, which provides an effective way to reduce redundant information in high-dimensional monitoring data. Second, it establishes a hybrid prediction framework that combines Tucker decomposition and LSTM in a unified pipeline for full-lifecycle EV battery analysis. This design is motivated by broader advances in low-rank tensor learning, latent factor analysis, and temporal representation modeling for complex structured data \cite{r1,r4,r8,r15,e18,e22,r28}.

\section*{2 Related Work}

Research on battery state prediction has increasingly shifted from conventional model-based estimation to data-driven learning frameworks. In battery-related applications, deep architectures have been used for remaining useful life prediction, capacity prediction, state monitoring, and embedded real-time inference, showing strong ability to capture nonlinear electrochemical behavior from measured data \cite{a1,a3,a5}. In particular, recurrent neural networks and LSTM-based models have become widely adopted because they are suitable for sequential dependence modeling in multivariate battery time series \cite{a7}. Nevertheless, purely sequential models may still face difficulties when the input space is highly redundant or contains multi-source couplings that are not explicitly organized before temporal learning \cite{a6}.

To enhance representation quality in high-dimensional environments, tensor decomposition and latent factor models have been extensively studied in the machine learning literature. Early tensor factor analysis provided an important mathematical foundation for representing multi-way data with compact latent structures \cite{a8}. Building on this idea, subsequent studies developed neural and regularized tensor factorization frameworks for incomplete, sparse, or nonlinear data representation, showing that low-rank tensor modeling can effectively preserve structural information while suppressing redundancy \cite{r1,r9,r11,r15,r28,r45}. These properties are highly relevant to battery monitoring data, which naturally exhibit multi-dimensional organization across variables, cycles, and time.

A substantial body of work has also explored latent factor analysis and tensor learning for dynamic prediction tasks. For example, temporal pattern-aware tensor factorization methods have been proposed for dynamic QoS estimation and structured forecasting, demonstrating the value of decomposing complex temporal observations into compact latent components \cite{a11,a12,e11,r10,r13,r17,r20,r21,r38}. Related studies further introduced adaptive, biased, and regularized tensor learning schemes to improve robustness in dynamically changing data environments \cite{r5,r6,r12,r16,r18,r19}. Although these methods were not designed specifically for battery SOC estimation, they provide useful methodological support for handling structured temporal data with hidden low-rank dependencies.

Recent research has further expanded tensor modeling toward neural, attention-based, and spatiotemporal representation learning. Neural Tucker factorization, mode-aware Tucker networks, tensor causal convolution, and self-attending tensor architectures have shown strong capability in capturing nonlinear interactions among multiple modes of data \cite{r4,r8,e18}. In addition, multi-indicator tensor recovery and robust latent feature analysis methods have demonstrated effectiveness in extracting stable representations from spatiotemporal signals \cite{e22,r40}. Such advances suggest that tensor-based models are not only suitable for compression, but also valuable for improving predictive expressiveness in downstream tasks.

Beyond tensor decomposition itself, a growing literature has investigated graph-enhanced, convolution-incorporated, and neighborhood-aware representation learning for complex relational and structured data. Representative examples include graph tensor convolutional networks, graph linear convolution pooling, adaptive neighborhood graph convolution, and high-order graph convolution with time-frequency transforms \cite{e4,e8,e24,r27,r36}. These studies collectively indicate that preserving structural correlation is crucial for robust learning in high-dimensional data spaces. For EV battery applications, this insight supports the use of decomposition-based preprocessing to expose latent interactions before sequence modeling.

Another closely related direction is low-rank completion and recovery for structured observations. High-order tensor completion, temporal-constraint tensor imputation, and robust low-rank recovery methods have been successfully used in traffic prediction, signal recovery, and visual data analysis \cite{r22,r33,r40}. These works reinforce the idea that compact latent structures can effectively characterize noisy and partially redundant measurements, which is particularly important for battery datasets collected under diverse real-world operating conditions.

Latent factor learning has also been improved through various optimization and regularization mechanisms. Existing studies have incorporated proximal ADMM, PID control, Kalman filtering, particle swarm optimization, and generalized accelerated gradient schemes into latent factor analysis in order to improve convergence stability and representation precision \cite{e2,e13,r3,r34,r41,r42,r43,r46}. Other works have considered feature weighting, diversified regularization, and nonnegative constraints to better model heterogeneous high-dimensional data \cite{r9,r11,r17,r39}. Although these techniques arise from different application scenarios, they collectively highlight the flexibility of latent factor models for extracting informative compact features.

From the application perspective, tensor and latent representation learning have already been applied to dynamic networks, traffic prediction, communication embedding, anomaly detection, electricity theft detection, recommendation, biomedical discovery, and healthcare analytics \cite{e1,e3,e5,e20,e21,e23,b3,b4,r23,r25,r26,r30,r31,r32}. Their success across these areas suggests that multi-way representation learning is broadly effective when the data involve temporal variation, hidden coupling, and high-dimensional redundancy. This cross-domain evidence provides further motivation for introducing Tucker decomposition into EV battery SOC prediction.

Overall, prior studies indicate two complementary trends. On the one hand, battery prediction research increasingly relies on learning-based temporal models to capture nonlinear dynamics \cite{a1,a3,a6,a7}. On the other hand, tensor-based and latent-factor approaches offer effective tools for compact representation, structural preservation, and robust learning in multi-dimensional data environments \cite{r1,r4,r15,e18,e22,r33}. However, relatively limited work has explicitly combined Tucker-style tensor compression with LSTM for SOC prediction using full-lifecycle EV operational data. The proposed Tucker-LSTM framework is intended to bridge this gap by integrating structured feature compression with temporal sequence modeling in a unified predictive architecture.

\section*{3 Main Work}

Inspired by recent developments in battery-oriented tensorized modeling and neural Tucker representation learning \cite{a2,e18,r73}, the main work of this paper can be summarized in four aspects.

\textbf{(1) Data curation and preprocessing:} Raw EV battery operation data were converted into a structured time-series dataset through temporal alignment, operational-state encoding, and sliding-window-based feature construction, enabling the learning of battery dynamics under diverse operating conditions.

\textbf{(2) Tucker-based feature dimensionality reduction:} A three-way sample$\times$window$\times$feature tensor was constructed, and Tucker decomposition was applied to extract compact latent factors and a core tensor, so that cross-mode interactions could be preserved while redundant feature information was suppressed. This design is supported by recent progress in mode-aware and autoencoding Tucker learning \cite{r4,r28}.

\textbf{(3) Hybrid sequence prediction architecture:} A conventional LSTM model and a Tucker-LSTM model were developed under the same hyperparameter setting to ensure a fair comparison. By combining tensor-based structural compression with sequential deep learning, the proposed architecture follows an effective paradigm for high-dimensional temporal prediction \cite{r8,r26}.

\textbf{(4) Comparative performance evaluation:} Quantitative metrics and qualitative visual analyses were jointly employed to assess predictive accuracy, fitting behavior, and residual characteristics, thereby providing a comprehensive validation of the proposed SOC prediction framework.

\section*{4 Methods}

\subsection*{4.1 Experimental Data}

The dataset used in this study contains full-lifecycle operating records of an electric vehicle, covering an odometer range from 0 to 38,548.7 km. The data were collected according to the Chinese national standard GB/T 32960.3-2016, ensuring consistency in the acquisition and transmission of onboard vehicle information. This real-world dataset provides continuous observations of battery behavior under practical driving and charging conditions.

% Table 1 
\begin{table}[H]
\centering
\caption{Description of battery operational features}
\adjustbox{width=1\textwidth}{
\begin{tabular}{>{\centering\arraybackslash}l>{\centering\arraybackslash}l>{\centering\arraybackslash}l|>{\centering\arraybackslash}l>{\centering\arraybackslash}l>{\centering\arraybackslash}l}
\toprule
Feature & Description & Unit/Range & Feature & Description & Unit/Range \\
\midrule
TIME & Unix timestamp & BYTE[6] & SOC & State of charge & \% \\
CHARGE\_STATUS & Charging state indicator & \{0, 1\} & MAX\_CELL\_VOLT & Maximum individual cell voltage & V \\
SPEED & Vehicle speed & km/h & MIN\_CELL\_VOLT & Minimum individual cell voltage & V \\
SUM\_MILE\_AGE & Accumulated mileage & km & MAX\_TEMP & Maximum battery temperature & $^\circ$C \\
SUM\_VOLTAGE & Total pack voltage & V & MIN\_TEMP & Minimum battery temperature & $^\circ$C \\
SUM\_CURRENT & Total pack current & A & CELL\_VOLT\_DIFF & Cell voltage spread & V \\
\bottomrule
\end{tabular}
}
\end{table}

As listed in Table 1, the recorded variables include Unix time, charging status, vehicle speed, driving range, total voltage, total current, state of charge (SOC), maximum and minimum cell voltages, maximum and minimum battery temperatures, and inter-cell voltage differences. Such multivariate battery data exhibit strong temporal dependency, heterogeneous feature composition, and potential incompleteness, which are common characteristics of high-dimensional industrial monitoring data \cite{e6,r75,r44}.

Considering that real-world time-series data may contain irregular fluctuations, missing entries, and cross-channel coupling effects, appropriate preprocessing is necessary before predictive modeling. Previous studies have shown that preserving temporal structure and latent low-rank information is beneficial for downstream prediction and recovery tasks in complex multivariate datasets \cite{r22,r71,r83,b10,b15}. Therefore, this dataset is well suited for the proposed tensor-based SOC prediction framework.

\subsection*{4.2 Tucker Decomposition for Feature Dimensionality Reduction}
Tucker decomposition\cite{a8} decomposes a multi-way array into a core tensor and mode-specific factor matrices, and therefore is very naturally suited for a three-way tensor organized as samples $\times$ timesteps $\times$ features, since it can perform feature-space dimensionality reduction by projecting the original feature mode onto a subspace of reduced rank without destroying the underlying multi-linear structure of the data \cite{r73,r4,e18}.

The procedure can be unambiguously and concisely described as follows: the training data tensor $X_{\text{train}} \in \mathbb{R}^{(n_{\text{samples}} \times \text{window}_{\text{size}} \times n_{\text{features}})}$ is first matricized along the feature mode to obtain $X_{\text{train}_{2d}} \in \mathbb{R}^{(n_{\text{features}} \times (n_{\text{samples}} \cdot \text{window}_{\text{size}}))}$, then its covariance matrix $C = X_{\text{train}_{2d}} X_{\text{train}_{2d}}^T$ is computed and decomposed by eigenanalysis as per Equation (1), following the general idea of mode-wise latent factor extraction in tensor representation learning \cite{r1,r15,r45}.

\begin{equation}
C=U\Lambda U^{T}
\end{equation}

Retaining only the leading $n_{\text{components}}$ eigenvectors (with $n_{\text{components}}=10$ in this study) forms the projection matrix $U_{\text{reduced}}$. This matrix is then applied to all datasets as shown in Equation (2), after which the projected data are reshaped to reinstate the temporal dimension, yielding a tensor of size $(n_{\text{samples}} \times \text{window}_{\text{size}} \times n_{\text{components}})$. From the given formula it is clear and concise to write:

\begin{equation}
X_{\text{proj}}=X_{2d}U_{\text{reduced}}
\end{equation}

Because this method is computationally efficient, requiring only the $n_{\text{features}} \times n_{\text{features}}$ covariance matrix rather than operating on the full tensor, and produces an orthogonal projection that maximizes retained variance in the feature space, it can be naturally regarded as principal component analysis applied to the unfolded tensor \cite{b3,e22,r33,r40}.

\subsection*{4.3 LSTM Network Architecture}
Since the LSTM \cite{a7} architecture deals with the input sequence $X = [x_1, x_2, \dots, x_T]$ by means of successive recurrence relations for hidden states $h_t$ and cell states $c_t$, the equations are given as follows. This recurrent design is well suited for modeling temporally dependent multivariate data, and remains a widely adopted sequence-learning mechanism when long-range temporal patterns need to be preserved \cite{r7,e4}:

\begin{equation}
\left\{
\begin{aligned}
i_t &= \sigma(W_i x_t + U_i h_{t-1} + b_i) \\
f_t &= \sigma(W_f x_t + U_f h_{t-1} + b_f) \\
o_t &= \sigma(W_o x_t + U_o h_{t-1} + b_o) \\
c_t &= f_t \odot c_{t-1} + i_t \odot \tanh(W_c x_t + U_c h_{t-1} + b_c) \\
h_t &= o_t \odot \tanh(c_t)
\end{aligned}
\right.
\end{equation}

where $i,f,o$ represent input, forget, and output gates, $\sigma$ denotes the sigmoid function, and $\odot$ element-wise multiplication. Similar sequence-aware and structure-preserving neural architectures have also been explored in graph- and tensor-oriented temporal learning tasks, further highlighting the importance of capturing both inter-variable dependency and temporal evolution in predictive modeling \cite{e8,r53,r56}.

The final prediction is obtained from the last hidden state through a fully connected layer:

\begin{equation}
y=W_{\text{fc}}h_T + b_{\text{fc}}
\end{equation}

Both models (standard LSTM and Tucker-LSTM) use identical architecture: 2 LSTM layers with hidden size 64, dropout 0.2 between layers, and a linear output layer. The only difference lies in input dimensionality: 14 features for standard LSTM versus 10 components for Tucker-LSTM. Such a controlled design helps isolate the effect of Tucker-based input compression from that of the sequence predictor itself, thereby enabling a fair comparison between original-space and reduced-space temporal modeling.

\subsection*{4.4 Training Protocol}
Since the data was divided chronologically with 70\% for training, 15\% for validation, and 15\% for testing, the temporal order was preserved and hence look-ahead bias was avoided. Input sequences were therefore formed by means of sliding windows of length 10 and stride 1. This temporally ordered splitting strategy is consistent with common practices in sequential prediction and time-dependent tensor modeling, where preserving causal structure is essential for reliable evaluation \cite{r31,r74}.

Training utilized the Adam optimizer having an initial learning rate of 0.01, a batch size of 128 and Mean Squared Error (MSE) loss. Learning rate decrease when on plateau (factor 0.5, patience set to 5 epochs). The validation loss served as the criterion for early stopping. Training terminated automatically when no improvement occurred for 15 consecutive epochs. Such an optimization setup is broadly compatible with modern deep sequence-learning and low-rank tensor learning frameworks, where adaptive gradient updates, validation-guided convergence control, and efficient implementation are commonly adopted to improve training stability and generalization \cite{r12,r58,r65}. All model implementations were conducted in PyTorch, with GPU-accelerated training employed whenever available, in line with current practice for computationally intensive neural and tensor-based modeling pipelines \cite{r80}.

\subsection*{4.5 Evaluation Metrics}
Model performance was quantified using complementary metrics including Mean Squared Error (MSE), Root Mean Squared Error (RMSE), Mean Absolute Error (MAE), and Coefficient of Determination ($R^2$), which respectively penalize large deviations, reflect average prediction error magnitude, and measure the proportion of variance explained by the model. The combined use of these metrics provides a more comprehensive assessment of regression performance and has been widely adopted in predictive modeling and data reconstruction studies \cite{b14,e11,r57}.

\section*{5 Results}
Table 2 and Figure 1 show that Tucker-LSTM outperforms standard LSTM across all metrics. On the test set, Tucker-LSTM gives an MSE of 6.22 (RMSE:2.49\%, MAE:1.73\%, $R^2$:0.976 ), while the baseline has an MSE of 21.07 ( RMSE:4.59\%, MAE:3.37\%, $R^2$:0.918 ), hence there is a 70.5\% reduction in MSE and a 48.7\% improvement in MAE.

% figure 1
\begin{figure}[H]
\centering
\includegraphics[width=1\textwidth]{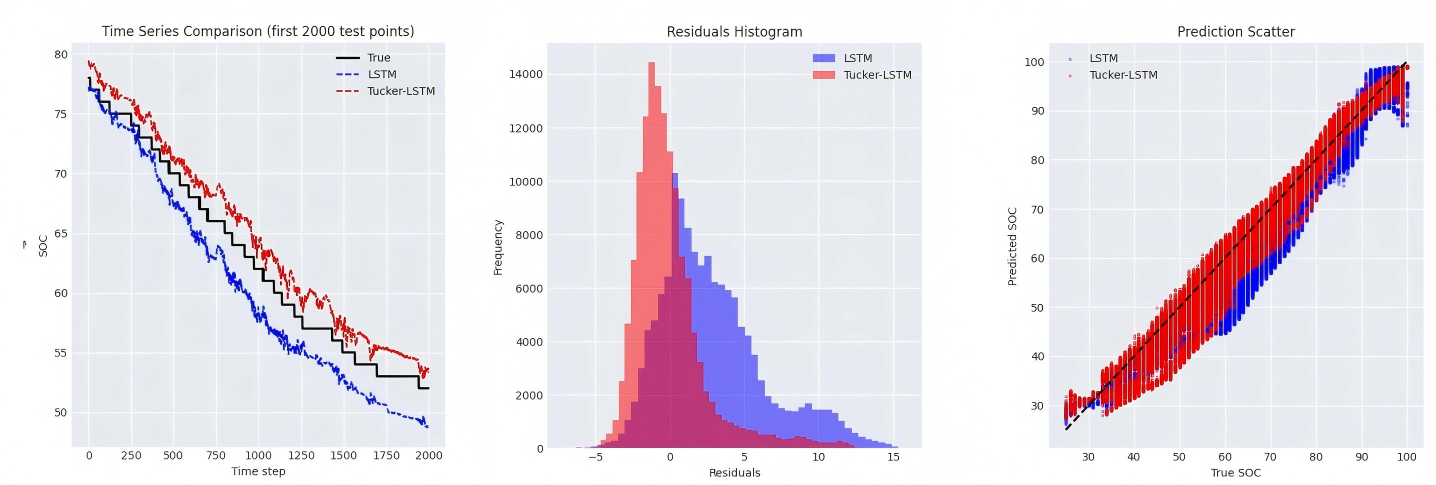}
\caption{Comparison chart of SOC predictions between LSTM and Tucker-LSTM}
\end{figure}

Fig. 1 contrasts the SOC estimation performance between LSTM and Tucker-LSTM. The time-series graph indicates that both models track the overall trend well, yet Tucker-LSTM matches the true values more precisely, particularly when there are swift SOC variations. The remaining histogram shows that Tucker-LSTM errors are more closely centered around zero with a steeper peak, suggesting smaller bias and variance. The dispersion plot verifies that Tucker-LSTM forecasts cluster more closely along the diagonal. In general, Tucker-LSTM shows better accuracy and stability, which is in line with the quantitative outcomes in Table 2.

% Table 2
\begin{table}[H]
\centering
\caption{Comparison experimental results of LSTM and Tucker-LSTM}
\begin{tabular}{lcccc}
\toprule
Model & MSE & RMSE(\%) & MAE (\%) & $R^2$ \\
\midrule
LSTM & 21.0735 & 4.5906 & 3.3725 & 0.9183 \\
Tucker-LSTM & 6.2179 & 2.4936 & 1.7286 & 0.9759 \\
\bottomrule
\end{tabular}
\end{table}

\section*{6 Conclusion}
This paper presents a Tucker decomposition-enhanced LSTM architecture for full-lifecycle SOC estimation in electric vehicles, and demonstrably shows that the method reduces feature dimensionality without destroying the temporal structure of the battery data, hence it outperforms the standard LSTM by 70.5\% in MSE and 48.7\% in MAE. Therefore, a natural and convincing conclusion is drawn: tensor decomposition effectively compresses high-dimensional battery data without compromising predictive accuracy. This finding is consistent with prior studies showing that tensor-based low-rank representation and structured latent feature extraction can preserve essential information while improving learning efficiency and predictive robustness in high-dimensional data analysis \cite{a8,r73,r4,e22,r40}.

Future research will incorporate other mainstream dimensionality reduction methods for comparative experiments to comprehensively validate Tucker's advantages across various scenarios, while further exploring the application of tensor analysis integrated with electric vehicle battery data. In particular, subsequent work may investigate alternative tensor factorization and representation-learning strategies, such as neural Tucker variants, low-rank tensor recovery, and graph-regularized latent modeling, in order to better characterize nonlinear dependency, temporal dynamics, and structured interactions in battery-related data \cite{e18,r33,r45,r57,r80}.

\bibliographystyle{IEEETran}
\bibliography{mybib}

@article{a1,
  title={A Lithium-Ion Battery Remaining Useful Life Prediction Model Based on CEEMDAN Data Preprocessing and HSSA-LSTM-TCN},
  author={ Qiu, Shaoming  and  Zhang, Bo  and  Lv, Yana  and  Zhang, Jie  and  Zhang, Chao },
  journal={World Electric Vehicle Journal},
  volume={15},
  number={5},
  year={2024},
}

@article{a2,
  title={Explainable real-time data driven method for battery electric model reconstruction via tensor train decomposition},
  author={ Ryzhov, A.  and  Rajinovic, K.  and  Kuhnelt, H.  and  De, Gennaro M. },
  journal={Journal of Power Sources},
  number={Jan.1},
  pages={625},
  year={2025},
}

@article{a3,
  title={Mechanistically guided residual learning for battery state monitoring throughout life},
  author={ Che, Yunhong  and  Zheng, Yusheng  and  Rhyu, Jinwook  and  Guo, Jia  and  Wang, Shimin  and  Teodorescu, Remus  and  Braatz, Richard D.  and  Bot, Nathalie Le  and  Bergin, Enda  and  Gillespie, Fiona },
  journal={Nature Communications},
  year={2026},
}

@article{a4,
  title={Method—Research on Multi-Scale Signal Processing and Efficient Model Construction Strategies in Lithium-Ion Battery State Prediction},
  author={ Gao, Zhijun  and  Dai, Rui  and  Ning, Yi  and  Guo, Xifeng },
  journal={Journal of The Electrochemical Society},
  volume={171},
  number={12},
  pages={120509},
  year={2024},
}

@article{a5,
  title={TensorRT Powered Model for Ultra-Fast Li-Ion Battery Capacity Prediction on Embedded Devices},
  author={ Zhu, Chunxiang  and  Qian, Jiacheng  and  Gao, Mingyu },
  journal={Energies (19961073)},
  volume={17},
  number={12},
  year={2024},
}

@article{a6,
  title={Latent-Factorization-of-Tensors-Incorporated Battery Cycle Life Prediction},
  author={ Chen, Minzhi  and  Tao, Li  and  Lou, Jungang  and  Luo, Xin },
  journal={IEEE/CAA Journal of Automatica Sinica},
  volume={12},
  number={3},
  pages={633-635},
  year={2025},
}

@article{a7,
  title={Long Short-Term Memory},
  author={ Hochreiter, S  and  Schmidhuber, J },
  journal={Neural Computation},
  volume={9},
  number={8},
  pages={1735-1780},
  year={1997},
}

@article{a8,
  title={Some mathematical notes on three-mode factor analysis},
  author={ Tucker, Ledyard },
  journal={Psychometrika},
  volume={31},
  number={3},
  pages={279-311},
  year={1966},
}

@article{a9,
    title = {An overview of data-driven battery health estimation technology for battery management system},
    author = {Minzhi Chen and Guijun Ma and Weibo Liu and Nianyin Zeng and Xin Luo},
    journal = {Neurocomputing},
    volume = {532},
    pages = {152-169},
    year = {2023}
}

@INPROCEEDINGS{a10,
    title={A Breif Review on Data-driven Battery Health Estimation Methods for Energy Storage Systems}, 
    author={Chen, Minzhi and Wu, Hao},
    booktitle={2022 IEEE International Conference on Networking, Sensing and Control (ICNSC)}, 
    pages={1-6},
    year={2022}
}

@ARTICLE{a11,
    title={A Generalized Nesterov's Accelerated Gradient-Incorporated Non-Negative Latent-Factorization-of-Tensors Model for Efficient Representation to Dynamic QoS Data}, 
    author={Chen, Minzhi and Wang, Renfang and Qiao, Yan and Luo, Xin},
    journal={IEEE Transactions on Emerging Topics in Computational Intelligence}, 
    volume={8},
    number={3},
    pages={2386-2400},
    year={2024}
}

@article{a12,
    title = {A momentum-incorporated latent factorization of tensors model for temporal-aware QoS missing data prediction},
    author = {Qingxian Wang and Minzhi Chen and Mingsheng Shang and Xin Luo},    
    journal = {Neurocomputing},
    volume = {367},
    pages = {299-307},
    year = {2019}
}

@Article{a13,
    title = {Analysis of State-of-Charge Estimation Methods for Li-Ion Batteries Considering Wide Temperature Range},
    author = {Miao, Yu and Gao, Yang and Liu, Xinyue and Liang, Yuan and Liu, Lin},
    journal = {Energies},
    volume = {18},
    NUMBER = {5},
    year = {2025},
    ARTICLE-NUMBER = {1188}
}

@Article{a14,
    TITLE = {Review of State-of-Charge Estimation Methods for Electric Vehicle Applications},
    AUTHOR = {Pisani Orta, Miguel Antonio and García Elvira, David and Valderrama Blaví, Hugo},
    JOURNAL = {World Electric Vehicle Journal},
    VOLUME = {16},
    YEAR = {2025},
    NUMBER = {2},
    ARTICLE-NUMBER = {87},
}

@Article{a15,
TITLE = {A Critical Review of the State Estimation Methods of Power Batteries for Electric Vehicles},
AUTHOR = {Zhang, Qi and Rong, Hailin and Zhao, Daduan and Pei, Menglu and Dong, Xing},
JOURNAL = {Energies},
VOLUME = {18},
YEAR = {2025},
NUMBER = {14},
ARTICLE-NUMBER = {3834}
}

@Article{a16,
TITLE = {Research Progress on State of Charge Estimation Methods for Power Batteries in New Energy Intelligent Connected Vehicles},
AUTHOR = {Li, Hongzhao and Jia, Hongsheng and Xiao, Ping and Jiang, Haojie and Chen, Yang},
JOURNAL = {Energies},
VOLUME = {18},
YEAR = {2025},
NUMBER = {9},
ARTICLE-NUMBER = {2144}
}

@Article{a17,
AUTHOR = {Pisani Orta, Miguel Antonio and García Elvira, David and Valderrama Blaví, Hugo},
TITLE = {Review of State-of-Charge Estimation Methods for Electric Vehicle Applications},
JOURNAL = {World Electric Vehicle Journal},
VOLUME = {16},
YEAR = {2025},
NUMBER = {2},
ARTICLE-NUMBER = {87}
}

@patent{a18,
  author = {Edison, Thomas A.},
  title  = {Electric Meter},
  year   = {1880},
  month  = jan,
  nationality = {United States},
  number = {US 223,898},
  note   = {Filed January 27, 1880, issued January 27, 1881}
}

@article{a19,
  author  = {Pop, V. and Bergveld, H. J. and Notten, P. H. L. and others},
  title   = {State-of-the-art of battery state-of-charge determination},
  journal = {Measurement Science and Technology},
  year    = {2005},
  volume  = {16},
  number  = {12},
  pages   = {R93--R110},
  doi     = {10.1088/0957-0233/16/12/R01}
}

@article{a20,
  author  = {Kalman, R. E.},
  title   = {A new approach to linear filtering and prediction problems},
  journal = {Journal of Basic Engineering},
  year    = {1960},
  volume  = {82},
  number  = {1},
  pages   = {35--45}
}

@article{a21,
  author  = {Plett, Gregory L.},
  title   = {Extended Kalman filtering for battery management systems of LiPB-based HEV battery packs. {Part} 2. {Modeling} and identification},
  journal = {Journal of Power Sources},
  year    = {2004},
  volume  = {134},
  number  = {2},
  pages   = {262--276},
  doi     = {10.1016/j.jpowsour.2004.02.032}
}

@ARTICLE{a22,
    title={Data-Driven Approaches for Estimation of EV Battery SoC and SoH: A Review}, 
    author={Padder, Shahid Gulzar and Ambulkar, Jayesh and Banotra, Atul and Modem, Sudhakar and Maheshwari, Sidharth and Jayaramulu, Kolleboyina and Kundu, Chinmoy},
    journal={IEEE Access}, 
    year={2025},
    volume={13},
    pages={35048-35067}}

@article{a23,
    title = {Electrochemical-Model-Based Estimation of Li-ion Battery SoC: Performance and Limitations of EKF Method in an Electromobility Use Case},
    author = {Shahbaz Ali and Antoneta Iuliana Bratcu and Pierre-Xavier Thivel and Iulian Munteanu},
    journal = {IFAC-PapersOnLine},
    volume = {59},
    number = {9},
    pages = {115-120},
    year = {2025},
    issn = {2405-8963},
}

@techreport{a24,
  author      = {Secunde, Richard R. and Birchenough, Arthur G.},
  title       = {Mercury electrochemical coulometer as a battery state-of-charge indicator},
  institution = {National Aeronautics and Space Administration (NASA)},
  year        = {1970},
  address     = {Washington, D.C.}
}

@ARTICLE{e1,
  title={Federated Latent Factorization of Tensors for Privacy-Preserving Representation Learning to Large-scale Dynamic Weighted Directed Network}, 
  author={Wu, Di and Zhong, Shuai and He, Yi and Luo, Xin and Gao, Xinbo},
  journal={IEEE Transactions on Dependable and Secure Computing}, 
  pages={1-16},
  year={2026},
  publisher={IEEE}
}

@ARTICLE{e2,
  title={Adaptive PID-Incorporated Nonnegative Latent Factor Analysis}, 
  author={Li, Jinli and Yuan, Ye and He, Tiantian and Luo, Xin},
  journal={IEEE Transactions on Systems, Man, and Cybernetics: Systems}, 
  pages={1-14},
  year={2026},
  publisher={IEEE}
}

@ARTICLE{e3,
  title={A Robust Approach to Electricity Theft Detection via Tensor Representation-Driven Contrastive Distillation}, 
  author={Qin, Wen and Ding, Yuting and Luo, Xin},
  journal={IEEE Transactions on Industrial Informatics}, 
  pages={1-10},
  year={2026},
  publisher={IEEE}
}

@ARTICLE{e4,
  title={Advanced High-Order Graph Convolutional Networks with Assorted Time-Frequency Transforms}, 
  author={Wang, Ling and Yuan, Ye and Luo, Xin},
  journal={IEEE/CAA Journal of Automatica Sinica}, 
  volume={13},
  number={2},
  pages={394-408},
  year={2026},
  publisher={IEEE}
}

@ARTICLE{e5,
  title={TraceHG: An Unsupervised Dual-View Framework for Microservice Anomaly Detection}, 
  author={Han, Ningning and Lu, Siyang and Lin, Zaichao and Li, Bin and Wang, Nan and Luo, Xin},
  journal={IEEE Transactions on Services Computing}, 
  volume={19},
  number={2},
  pages={1633-1646},
  year={2026},
  publisher={IEEE}
}

@ARTICLE{e6,
  title={Multimetric Autoencoder for Representing High-Dimensional and Incomplete Data}, 
  author={Wu, Di and Liang, Cheng and He, Yi and Qiao, Yan and Luo, Xin},
  journal={IEEE Transactions on Systems, Man, and Cybernetics: Systems}, 
  volume={56},
  number={3},
  pages={1533-1546},
  year={2026},
  publisher={IEEE}
}

@ARTICLE{e8,
  title={Graph Tensor Convolutional Network}, 
  author={Wang, Ling and Yuan, Ye and Luo, Xin},
  journal={IEEE Transactions on Systems, Man, and Cybernetics: Systems}, 
  volume={56},
  number={5},
  pages={3008-3024},
  year={2026},
  publisher={IEEE}
}

@ARTICLE{e11,
  title={A Sampling-Neighborhood-Regularized Latent Factorization of Tensor for Dynamic QoS Estimation}, 
  author={Xu, Xiuqin and Lin, Mingwei and Xu, Zeshui and Luo, Xin},
  journal={IEEE Transactions on Network and Service Management}, 
  volume={23},
  pages={1707-1722},
  year={2026},
  publisher={IEEE}
}

@ARTICLE{e13,
  title={Genetic Algorithm-Based Two-Step Optimization for Precise Latent Factor Analysis}, 
  author={Lyu, Chao and Cheng, Jingna and Luo, Xin and Shi, Yuhui},
  journal={IEEE Transactions on Neural Networks and Learning Systems}, 
  volume={37},
  number={5},
  pages={2294-2306},
  year={2026},
  publisher={IEEE}
}

@ARTICLE{e18,
  title={Multi-Aspect Self-Attending Neural Tucker Factorization for Spatiotemporal Representation Learning}, 
  author={Hou, Yikai and Tang, Peng and Luo, Xin},
  journal={IEEE/CAA Journal of Automatica Sinica}, 
  volume={13},
  number={4},
  pages={986-988},
  year={2026},
  publisher={IEEE}
}

@INPROCEEDINGS{e20,
  title={Federated Deep Latent Factor Model for Privacy-Preserving Recommendation}, 
  author={Gao, JunXiang and Wu, Di and Chen, Jia and Zhou, Min and Luo, Xin},
  booktitle={2025 IEEE International Conference on Systems, Man, and Cybernetics (SMC)}, 
  pages={1689-1694},
  year={2025},
  publisher={IEEE}
}

@INPROCEEDINGS{e21,
  title={Adversarial Domain Adaptation for Accurately Predicting the Associations between Herbal Compounds and Target Proteins}, 
  author={Qiao, Yantong and Hu, Lun and Zhang, Jun and Hu, Pengwei and Luo, Xin},
  booktitle={2025 IEEE International Conference on Systems, Man, and Cybernetics (SMC)}, 
  pages={3072-3077},
  year={2025},
  publisher={IEEE}
}

@INPROCEEDINGS{e22,
  title={Multi-Indicator Latent Factorization of Tensors for Spatio-Temporal Signal Recovery}, 
  author={Yu, Chengjun and Wu, Di and Chen, Jia and Zhou, Min and Luo, Xin},
  booktitle={2025 IEEE 31th International Conference on Parallel and Distributed Systems (ICPADS)}, 
  pages={1-8},
  year={2025},
  publisher={IEEE}
}

@ARTICLE{e23,
  title={Advancing Healthcare with Large Language Models: Techniques and Application}, 
  author={Hu, Zhenlin and Peng, Zhizhi and Bi, Zhen and Shen, Qing and Liu, Zhenfang and Lou, Jungang and Luo, Xin},
  journal={IEEE/CAA Journal of Automatica Sinica}, 
  volume={12},
  number={12},
  pages={2371-2398},
  year={2025},
  publisher={IEEE}
}

@ARTICLE{e24,
  title={Graph Linear Convolution Pooling for Learning in Incomplete High-Dimensional Data}, 
  author={Bi, Fanghui and He, Tiantian and Ong, Yew-Soon and Luo, Xin},
  journal={IEEE Transactions on Knowledge and Data Engineering}, 
  volume={37},
  number={4},
  pages={1838-1852},
  year={2025},
  publisher={IEEE}
}

@article{r1,
  title={Neural Nonnegative Latent Factorization of Tensors Model With Acceleration and Unconstraint},
  author={Li, Wenqiang and Lin, Mingwei and Xu, Xiuqin and Lin, Ling and Xu, Zeshui and Luo, Xin},
  journal={IEEE Transactions on Systems, Man, and Cybernetics: Systems},
  volume={56},
  number={1},
  pages={164-178},
  year={2026},
  publisher={IEEE}
}

@article{r3,
  title={Dynamic Stochastic Reorientation Particle Swarm Optimization for Adaptive Latent Factor Analysis in High-Dimensional Sparse Matrices},
  author={Lyu, Chao and Ma, Ziwen and Luo, Xin and Shi, Yuhui},
  journal={IEEE Transactions on Knowledge and Data Engineering},
  volume={38},
  number={1},
  pages={222-234},
  year={2026},
  publisher={IEEE}
}

@article{r4,
  title={Learning Accurate Representation to Nonstandard Tensors via a Mode-Aware Tucker Network},
  author={Wu, Hao and Wang, Qu and Luo, Xin and Wang, Zidong},
  journal={IEEE Transactions on Knowledge and Data Engineering},
  volume={37},
  number={12},
  pages={7272-7285},
  year={2025},
  publisher={IEEE}
}

@article{r5,
  title={A Convolution Bias-Incorporated Nonnegative Latent Factorization of Tensors Model for Accurate Representation Learning to Dynamic Directed Graphs},
  author={Wang, Qu and Wu, Hao and Luo, Xin},
  journal={IEEE Transactions on Systems, Man, and Cybernetics: Systems},
  volume={55},
  number={12},
  pages={8902-8914},
  year={2025},
  publisher={IEEE}
}

@article{r6,
  title={A Proximal-ADMM-Incorporated Nonnegative Latent-Factorization-of-Tensors Model for Representing Dynamic Cryptocurrency Transaction Network},
  author={Liao, Xin and Wu, Hao and He, Tiantian and Luo, Xin},
  journal={IEEE Transactions on Systems, Man, and Cybernetics: Systems},
  volume={55},
  number={11},
  pages={8387-8401},
  year={2025},
  publisher={IEEE}
}

@article{r7,
  title={Discovering Spatiotemporal--Individual Coupled Features From Nonstandard Tensors—A Novel Dynamic Graph Mixer Approach},
  author={Bi, Fanghui and He, Tiantian and Ong, Yew-Soon and Luo, Xin},
  journal={IEEE Transactions on Neural Networks and Learning Systems},
  volume={36},
  number={11},
  pages={19834-19848},
  year={2025},
  publisher={IEEE}
}

@article{r8,
  title={A novel tensor causal convolution network model for highly-accurate representation to spatio-temporal data},
  author={Liao, Xin and Wu, Hao and Luo, Xin},
  journal={IEEE Transactions on Automation Science and Engineering},
  volume={22},
  pages={19525-19537},
  year={2025},
  publisher={IEEE}
}

@article{r9,
  title={A Fine-Grained Regularization Scheme for Nonnegative Latent Factorization of High-Dimensional and Incomplete Tensors},
  author={Wu, Hao and Qiao, Yan and Luo, Xin},
  journal={IEEE Trans. Serv. Comput.},
  volume={17},
  number={6},
  pages={3006--3021},
  year={2024},
  publisher={IEEE}
}

@article{r10,
  title={Temporal pattern-aware QoS prediction by biased non-negative Tucker factorization of tensors},
  author={Tang, Peng and Ruan, Tao and Wu, Hao and Luo, Xin},
  journal={Neurocomputing},
  volume={582},
  pages={127447},
  year={2024},
  publisher={Elsevier}
}

@article{r11,
  title={Advancing non-negative latent factorization of tensors with diversified regularization schemes},
  author={Wu, Hao and Luo, Xin and Zhou, MengChu},
  journal={IEEE Transactions on Services Computing},
  volume={15},
  number={3},
  pages={1334-1344},
  year={2022},
  publisher={IEEE}
}

@article{r12,
  title={A PID-incorporated latent factorization of tensors approach to dynamically weighted directed network analysis},
  author={Wu, Hao and Luo, Xin and Zhou, MengChu and Rawa, Muhyaddin J and Sedraoui, Khaled and Albeshri, Aiiad},
  journal={IEEE/CAA Journal of Automatica Sinica},
  volume={9},
  number={3},
  pages={533-546},
  year={2022},
  publisher={IEEE}
}

@article{r13,
  title={Temporal pattern-aware QoS prediction via biased non-negative latent factorization of tensors},
  author={Luo, Xin and Wu, Hao and Yuan, Huaqiang and Zhou, MengChu},
  journal={IEEE transactions on cybernetics},
  volume={50},
  number={5},
  pages={1798-1809},
  year={2020},
  publisher={IEEE}
}

@article{r15,
  title={NeuLFT: A novel approach to nonlinear canonical polyadic decomposition on high-dimensional incomplete tensors},
  author={Luo, Xin and Wu, Hao and Li, Zechao},
  journal={IEEE Transactions on Knowledge and Data Engineering},
  volume={35},
  number={6},
  pages={6148-6166},
  year={2023},
  publisher={IEEE}
}

@inproceedings{r16,
  title={Neural latent factorization of tensors for dynamically weighted directed networks analysis},
  author={Wu, Hao and Luo, Xin and Zhou, MengChu},
  booktitle={2021 IEEE International Conference on Systems, Man, and Cybernetics (SMC)},
  pages={3061-3066},
  year={2021},
  organization={IEEE}
}

@inproceedings{r17,
  title={Instance-frequency-weighted regularized, nonnegative and adaptive latent factorization of tensors for dynamic QoS analysis},
  author={Wu, Hao and Luo, Xin},
  booktitle={2021 IEEE International Conference on Web Services (ICWS)},
  pages={560-568},
  year={2021},
  organization={IEEE}
}

@inproceedings{r18,
  title={Discovering hidden pattern in large-scale dynamically weighted directed network via latent factorization of tensors},
  author={Wu, Hao and Luo, Xin and Zhou, MengChu},
  booktitle={2021 IEEE 17th International Conference on Automation Science and Engineering (CASE)},
  pages={1533-1538},
  year={2021}
}

@inproceedings{r19,
  title={Proportional-integral-derivative-incorporated latent factorization of tensors for large-scale dynamic network analysis},
  author={Wu, Hao and Xia, Yan and Luo, Xin},
  booktitle={2021 China Automation Congress (CAC)},
  pages={2980-2984},
  year={2021},
  organization={IEEE}
}

@inproceedings{r20,
  title={Efficient representation to dynamic QoS data via generalized nesterov’s accelerated gradient-incorporated biased non-negative latent factorization of tensors},
  author={Chen, Minzhi and Luo, Xin},
  booktitle={2021 IEEE International Conference on Systems, Man, and Cybernetics (SMC)},
  pages={576-581},
  year={2021},
  organization={IEEE}
}

@article{r21,
  title={Adjusting learning depth in nonnegative latent factorization of tensors for accurately modeling temporal patterns in dynamic QoS data},
  author={Luo, Xin and Chen, Minzhi and Wu, Hao and Liu, Zhigang and Yuan, Huaqiang and Zhou, MengChu},
  journal={IEEE Transactions on Automation Science and Engineering},
  volume={18},
  number={4},
  pages={2142-2155},
  year={2021},
  publisher={IEEE}
}

@article{r22,
  author={Chen, Hong and Lin, Mingwei and Zhao, Liang and Xu, Zeshui and Luo, Xin},
  journal={IEEE Transactions on Intelligent Transportation Systems}, 
  title={Fourth-Order Dimension Preserved Tensor Completion With Temporal Constraint for Missing Traffic Data Imputation}, 
  volume={26},
  number={5},
  pages={6734-6748},
  year={2025},
  publisher={IEEE}
}

@book{r23,
  title={Dynamic Network Representation Based on Latent Factorization of Tensors},
  author={Wu, Hao and Wu, Xuke and Luo, Xin},
  year={2023},
  publisher={Springer}
}

@inproceedings{r25,
  title={Sgd-dyg: Self-reliant global dependency apprehending on dynamic graphs},
  author={Han, Minglian and Wang, Ling and Yuan, Ye and Luo, Xin},
  booktitle={Proceedings of the 31st ACM SIGKDD Conference on Knowledge Discovery and Data Mining V. 2},
  numpages={12},
  pages={802-813},
  year={2025},
  publisher={Association for Computing Machinery},
}

@article{b3,
  title={Biased Block Term Tensor Decomposition for Temporal Pattern-aware QoS Prediction},
  author={Wang, Qu and Liao, Xin and Wu, Hao},
  journal={International Journal of Pattern Recognition and Artificial Intelligence},
  volume = {39},
  number = {02},
  pages = {2550001},
  year={2025},
  publisher={World Scientific}
}

@inproceedings{b4,
  title={An Adaptive Temporal-Dependent Tensor Low-Rank Representation Model for Dynamic Communication Network Embedding},
  author={Liao, Xin and Hu, Qicong and Tang, Peng},
  booktitle={2024 International Conference on Networking, Sensing and Control (ICNSC)},
  pages={1-6},
  year={2024},
  organization={IEEE}
}

@ARTICLE{r26,
  title={A 3D Convolution-Incorporated Dimension Preserved Decomposition Model for Traffic Data Prediction}, 
  author={Lin, Mingwei and Liu, Jiaqi and Chen, Hong and Xu, Xiuqin and Luo, Xin and Xu, Zeshui},
  journal={IEEE Transactions on Intelligent Transportation Systems}, 
  volume={26},
  number={1},
  pages={673-690},
  year={2025},
  publisher={IEEE}
}

@article{r27,
  title={An Adaptive Neighborhood-Resonated Graph Convolution Network for Undirected Weighted Graph Representation},
  author={Chen, Jiufang and Yuan, Ye and Luo, Xin and Gao, Xinbo},
  journal={IEEE Transactions on Neural Networks and Learning Systems},
  volume={36},
  number={11},
  pages={19939-19950},
  year={2025},
  publisher={IEEE}
}

@article{r28,
  title={Auto-encoding neural tucker factorization},
  author={Tang, Peng and Luo, Xin and Woodcock, Jim},
  journal={IEEE Transactions on Knowledge and Data Engineering},
  volume={37},
  number={10},
  pages={5795-5807},
  year={2025},
  publisher={IEEE}
}

@article{r29,
  title={Neural Networks-Incorporated Latent Factor Analysis for High-Dimensional and Incomplete Data},
  author={Lin, Mingwei and Lin, Xingyu and Xu, Xiuqin and Xu, Zeshui and Luo, Xin},
  journal={IEEE Transactions on Systems, Man, and Cybernetics: Systems},
  volume={55},
  number={10},
  pages={7302-7314},
  year={2025},
  publisher={IEEE}
}

@article{r30,
  title={Identifying novel therapeutic targets of natural compounds in traditional Chinese medicine herbs with hypergraph representation learning},
  author={Qiao, Yantong and Hu, Lun and Zhang, Jun and Hu, Pengwei and Luo, Xin},
  journal={Briefings in Bioinformatics},
  volume={26},
  number={4},
  pages={bbaf399},
  year={2025},
  publisher={Oxford University Press}
}

@article{r31,
  title={FMvPCI: A Multiview Fusion Neural Network for Identifying Protein Complex via Fuzzy Clustering},
  author={Yang, Yue and Hu, Lun and Li, Guodong and Li, Dongxu and Hu, Pengwei and Luo, Xin},
  journal={IEEE Transactions on Systems, Man, and Cybernetics: Systems},
  volume={55},
  number={9},
  pages={6189-6202},
  year={2025},
  publisher={IEEE}
}

@article{r32,
  title={MNL: A highly-efficient model for large-scale dynamic weighted directed network representation},
  author={Chen, Minzhi and He, Chunlin and Luo, Xin},
  journal={IEEE Transactions on Big Data},
  volume={9},
  number={3},
  pages={889-903},
  year={2023},
  publisher={IEEE}
}

@article{r33,
  title={Low-rank high-order tensor completion with applications in visual data},
  author={Qin, Wenjin and Wang, Hailin and Zhang, Feng and Wang, Jianjun and Luo, Xin and Huang, Tingwen},
  journal={IEEE Transactions on Image Processing},
  volume={31},
  pages={2433--2448},
  year={2022},
  publisher={IEEE}
}

@article{r34,
  title={A proportional integral controller-enhanced non-negative latent factor analysis model},
  author={Yuan, Ye and Lu, Siyang and Luo, Xin},
  journal={IEEE/CAA Journal of Automatica Sinica},
  volume={12},
  number={6},
  pages={1246-1259},
  year={2025},
  publisher={IEEE}
}

@article{r36,
  title={Enhancing graph convolutional networks with an efficient k-hop neighborhood approach},
  author={Chen, Jiufang and Luo, Xin and Yuan, Ye and Wang, Zidong},
  journal={Information Fusion},
  volume = {124},
  pages = {103297},
  year={2025},
  issn = {1566-2535},
  publisher={Elsevier}
}

@article{r37,
  title={From data analysis to intelligent maintenance: a survey on visual defect detection in aero-engines},
  author={Wu, Peishu and Li, Han and Luo, Xin and Hu, Liwei and Yang, Rui and Zeng, Nianyin},
  journal={Measurement Science and Technology},
  volume={36},
  number={6},
  pages={062001},
  year={2025},
  publisher={IOP Publishing}
}

@article{r38,
  title={An Adaptively Bias-Extended Non-Negative Latent Factorization of Tensors Model for Accurately Representing the Dynamic QoS Data}, 
  author={Xu, Xiuqin and Lin, Mingwei and Luo, Xin and Xu, Zeshui},
  journal={IEEE Transactions on Services Computing}, 
  volume={18},
  number={2},
  pages={603-617},
  year={2025},
  publisher={IEEE}
}

@article{r39,
  title={Nonnegative Latent Factor Analysis-Incorporated and Feature-Weighted Fuzzy Double $ c $-Means Clustering for Incomplete Data},
  author={Song, Yan and Li, Ming and Zhu, Zhengyu and Yang, Guisong and Luo, Xin},
  journal={IEEE Transactions on Fuzzy Systems},
  volume={30},
  number={10},
  pages={4165-4176},
  year={2022},
  publisher={IEEE}
}

@article{r40,
  title={Robust Low-Rank Latent Feature Analysis for Spatiotemporal Signal Recovery},
  author={Wu, Di and Li, Zechao and Yu, Zhikai and He, Yi and Luo, Xin},
  journal={IEEE Transactions on Neural Networks and Learning Systems},
  volume={36},
  number={2},
  pages={2829-2842},
  year={2025},
  publisher={IEEE}
}

@article{r41,
  title={Proximal alternating-direction-method-of-multipliers-incorporated nonnegative latent factor analysis},
  author={Bi, Fanghui and Luo, Xin and Shen, Bo and Dong, Hongli and Wang, Zidong},
  journal={IEEE/CAA Journal of Automatica Sinica},
  volume={10},
  number={6},
  pages={1388-1406},
  year={2023},
  publisher={IEEE}
}

@article{r42,
  title={A Kalman-filter-incorporated latent factor analysis model for temporally dynamic sparse data},
  author={Yuan, Ye and Luo, Xin and Shang, Mingsheng and Wang, Zidong},
  journal={IEEE Transactions on Cybernetics},
  volume={53},
  number={9},
  pages={5788-5801},
  year={2023},
  publisher={IEEE}
}

\end{document}